\documentclass[journal]{IEEEtran}

\usepackage{times}
\usepackage{epsfig}
\usepackage{graphicx}
\usepackage{amsmath}
\usepackage{amssymb}
\usepackage{multirow}
\usepackage{booktabs}
\usepackage{float}
\usepackage{color}

\DeclareMathOperator*{\argmax}{argmax}

\begin{document}

\title{FSD-10: A Dataset for Competitive Sports Content Analysis}

\author{Shenglan~Liu~\IEEEmembership{Member,~IEEE},
        Xiang~Liu,
        Gao Huang,
        Lin Feng,
        Lianyu Hu,
        Dong Jiang,
        Aibin Zhang,
        Yang Liu,
        Hong Qiao~\IEEEmembership{Fellow,~IEEE}
\thanks{Shenglan Liu, Xiang Liu, Lin Feng, Lianyu Hu, Dong Jiang, Aibin Zhang, Yang Liu are with Faculty of Electronic Information and Electrical Engineering, Dalian University of Technology, Dalian, Liaoning, 116024 China. Email: ($\{$liusl, hly19961121, s201461147, 18247666912, dlut$\_$liuyang$\}$@mail.dlut.edu.cn, fenglin@dlut.edu.cn, xliudut@gmail.com).}
\thanks{Gao Huang is with Faculty of Tsinghua University of Technology, Beijing, China. Email: gaohuang@tsinghua.edu.cn.}
\thanks{H. Qiao is with the State Key Laboratory of Management and Control for Complex Systems, Institute of Automation, Chinese Academy of Sciences, Beijing 100190 China. E-mail: hong.qiao@ia.ac.cn.}
\thanks{This work is supported in part by The National Key Research and Development Program of China (2017YFB1300200)
National Natural Science Foundation of People's Republic of China (No. 61672130, 61602082, 91648205, 31871106),
the National Key Scientific Instrument and Equipment Development Project (No. 61627808),
the Development of Science and Technology of Guangdong Province Special Fund Project Grants (No. 2016B090910001),
the LiaoNing Revitalization Talents Program (No.XLYC1086006).}
        }

\markboth{Journal of \LaTeX\ Class Files,~Vol.~14, No.~8, August~2015}%
{Shell \MakeLowercase{\textit{et al.}}: Bare Demo of IEEEtran.cls for IEEE Journals}
\maketitle

\begin{abstract}
Action recognition is an important and challenging problem in video analysis.
Although the past decade has witnessed progress in action recognition with the development of deep learning, such process has been slow in competitive sports content analysis.
To promote the research on action recognition from competitive sports video clips, we introduce a Figure Skating Dataset (FSD-10) for finegrained sports content analysis.
To this end, we collect 1484 clips from the worldwide figure skating championships in 2017-2018, which consist of 10 different actions in men/ladies programs.
Each clip is at a rate of 30 frames per second with resolution 1080 $\times$ 720.
These clips are then annotated by experts in type, grade of execution, skater info, .etc.
To build a baseline for action recognition in figure skating, we evaluate state-of-the-art action recognition methods on FSD-10.
Motivated by the idea that domain knowledge is of great concern in sports field, we propose a keyframe based temporal segment network (KTSN) for classification and achieve remarkable performance.
Experimental results demonstrate that FSD-10 is an ideal dataset for benchmarking action recognition algorithms, as it requires to accurately extract action motions rather than action poses.
We hope FSD-10, which is designed to have a large collection of finegrained actions, can serve as a new challenge to develop more robust and advanced action recognition models.

\end{abstract}

\begin{figure*}[htbp]
\centering
\includegraphics[width=6.8in]{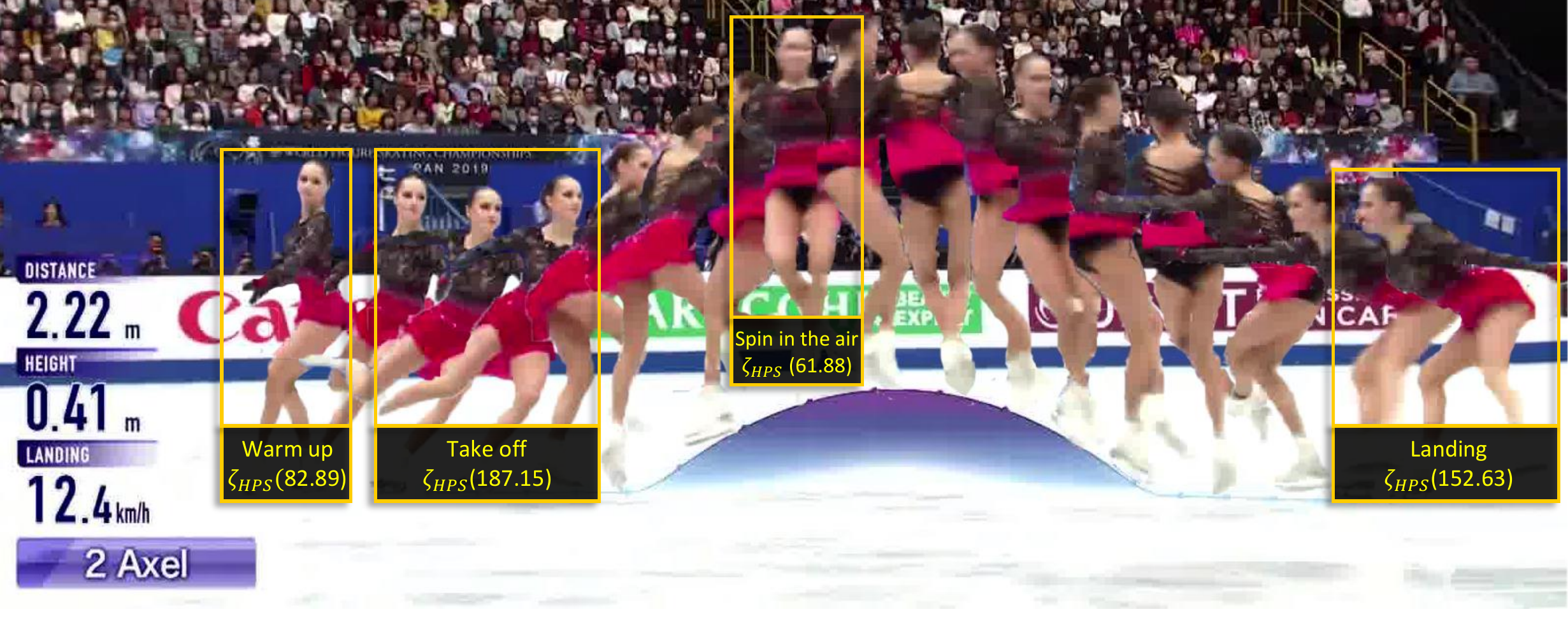}
\caption{An over view of the figure skating dataset. It is illustrated that warm up, take off, spinning in the air and landing are 4 steps of an Axel jump. $\zeta_{HPS}$ indicates the human pose scatter of a gesture, which is introduced in Sec \ref{HPS}.}
\label{fig_first}
\end{figure*}

\section{Introduction}

Due to the popularity of media-sharing platforms(e.g. YouTube), sports content analysis (SCA\cite{shih2017survey}) has become an important research topic in computer vision \cite{van2010race}\cite{gudmundsson2017spatio}\cite{safdarnejad2015sports}.
A vast amount of sports videos are piled up in computer storage, which are potential resources for deep learning.
In recent years, many enterprises (e.g. Bloomberg, SAP) have focus on SCA\cite{shih2017survey}.
In SCA, datasets are required to reflect characteristics of competitive sports, which is a guarantee for training deep learning models.
Generally, competitive sports content is a series of diversified, high professional and ultimate actions.
Unfortunately, existing trending human motion datasets (e.g. HMDB51\cite{kuehne2011hmdb}, UCF50\cite{soomro2012ucf101}) or action datasets of human sports (e.g. MIT Olympic sports\cite{pirsiavash2014assessing}, Nevada Olympic sports\cite{parmar2019action}) are not quite representative of the richness and complexity of competitive sports.
The limitations of current datasets could be summarised as follows:
(1) Current tasks of human action analysis are insufficient, which is limited to action types and contents of videos.
(2) In most classification tasks, discriminant of an action largely depends on both static human pose and background;
(3) Video segmentation, as an important task of video understanding (including classification and assessment in SCA ), is rarely discussed in current datasets.

\begin{table*}[htbp]
	\caption{A summary of existing action datasets}
	\setlength{\tabcolsep}{5mm}
	\label{table_dataset}
\begin{center}
\begin{tabular}{ccccccc}
	\toprule
	Dataset & Year & Actions & Clips & Classification & Score & Temporal Segmentation \\
	\midrule
    UCF Sports\cite{rodriguez2008action} & 2009 & 9 & 14-35 & $\surd$ & - & - \\
    Hollywood2\cite{marszalek2009actions} & 2009 & 12 & 61-278 & $\surd$ & - & - \\
    UCF50\cite{soomro2012ucf101} & 2010 & 50 & min. 100 & $\surd$ & - & - \\
    HMDB51\cite{kuehne2011hmdb} & 2011 & 51 & min. 101 & $\surd$ & - & - \\
    MIT Olympic sports\cite{pirsiavash2014assessing}  & 2015 & 2 & 150-159 & $\surd$ & $\surd$ & - \\
    Nevada Olympic sports\cite{parmar2019and} & 2017 & 3 & 150-370 & $\surd$ & $\surd$ & - \\
    AQA-7\cite{parmar2019action} & 2018 & 7 & 83-370 & $\surd$ & $\surd$ & - \\
    \textbf{FSD-10} & \textbf{2019} & \textbf{10} & \textbf{91-233} & \textbf{$\surd$} & \textbf{$\surd$} & \textbf{$\surd$} \\
	\bottomrule
\end{tabular}
\end{center}
\end{table*}

\begin{figure*}[htbp]
	\centerline{\includegraphics[width=7in]{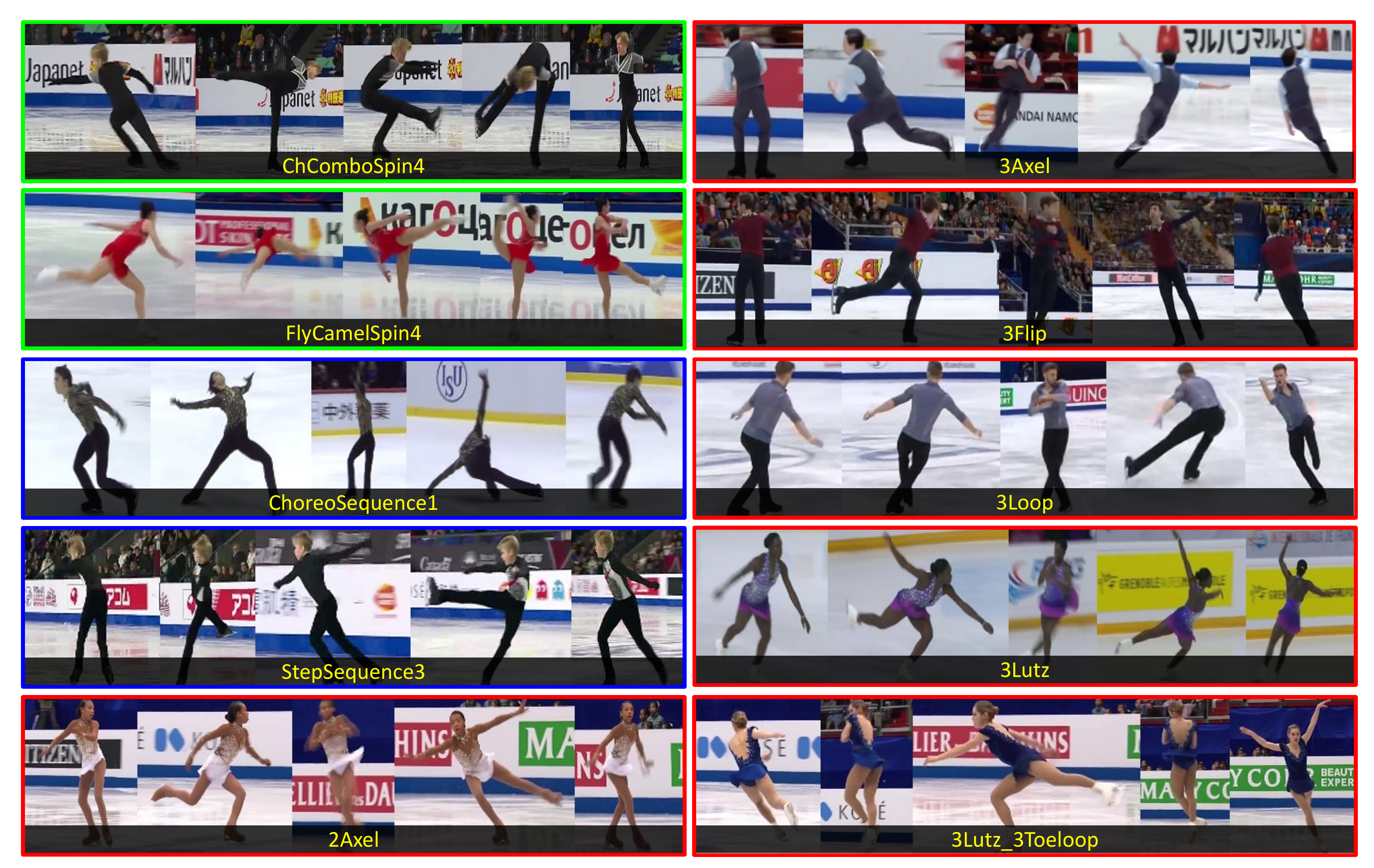}}
	\caption{10 figure skating actions included in FSD-10 shown with 5 frames. The color of frame borders specifies to which action category they belong (\textcolor[rgb]{0, 1, 0}{Spin}, \textcolor[rgb]{0, 0, 1}{Sequence}, \textcolor[rgb]{1, 0, 0}{Jump}) and the labels specifies their action types.}
	\label{fig_display}
\end{figure*}

To address the above issues, this paper proposes a figure skating dataset called FSD-10.
FSD-10 consists of 1484 figure skating videos with 10 different actions manually labeled.
These skating videos are segmented from around 80 hours of competitions of worldwide figure skating championships \footnote{ISU Grand Prix of Figure Skating, European Figure Skating Championships, Four Continents Figure Skating Championships (https://www.isu.org/figure-skating)} in 2017-2018.
FSD-10 videos range from 3 seconds to 30 seconds, and the camera is moving to focus the skater to ensure that he (she) appears in each frame during the process of actions.
Compared with existing datasets, our proposed dataset has several appealing properties.
First, actions in FSD-10 are complex in content and fast in speed.
For instance, the complex 2-loop-Axel jump in Fig. 1 is finished in only about 2s.
It's worth note that the jump type heavily depends on the take off process, which is a hard-captured moment.
Second, actions of FSD-10 are original from figure skating competitions, which are consistent in type and sports environment (only skater and ice around).
The above two aspects create difficulties for machine learning model to conclude the action types by a single pose or background.
Third, FSD-10 can be applied to standardized action quality assessment (AQA) tasks.
The AQA of FSD-10 actions depends on base value (BV) and grade of execution (GOE), which increases the difficulty of AQA by both action qualities and rules of International Skating Union (ISU)\footnote{One action appearing in second half of program will earn extra  rate of $10\%$ score of BV than that appearing in first half of program. This means that same action may get different BV score because of the rules.}.
Last, FSD-10 can be extended to more tasks than the mentioned datasets.

Along with the introduction of FSD-10, we introduce a key frame indicator called human pose scatter (HPS).
Based on HPS, we adopt key frame sampling to improve current video classification methods and evaluate these methods on FSD-10.
Furthermore, experimental results validate that key frame sampling is an important approach to improve performance of frame-based model in FSD-10, which is in concert with cognition rules of human in figure skating.
The main contributions of this paper can be summarised as follows.

\begin{itemize}
\item To our best knowledge, FSD-10 is the first challenging dataset with high-speed,  motion-focused and complex sport actions in competitive sports field, which introduces multiple tasks for computer vision, such as (fine grained) classification, long/short temporal human motion segmentation, action quality assessments, and program content assessment in time to music.
\item To set a baseline for future achievements, we also benchmark state-of-the-art sport classification methods on FSD-10. Besides, the key frame sampling is proposed to capture the pivotal action details in competitive sports, which achieves better performance than state-of-the-art methods in FSD-10.
\end{itemize}

In addition,compared to current datasets, we hope FSD-10 will be a standard dataset which promotes the research on background-independent action recognition, allowing models to be more robust to unexpected events
and generalize to novel situations and tasks.

\section{Related Works}

Current action datasets  could be divided into human action dataset (HAD) and human professional sports dataset (HPSD), which are listed in Tab. \ref{table_dataset}.
HAD consists of different kinds of broadly defined human motion clips, such as haircut and long jump.
For example, UCF101\cite{soomro2012ucf101} and HMDB\cite{kuehne2011hmdb} are typical cases of HAD, which have greatly promoted the process of video classification techniques.
In contrast, HPSD is a series of competitive sports actions, which is proposed for action quality assessment (AQA) tasks except classification.
MIT Olympic sports\cite{pirsiavash2014assessing} and Nevada Olympic sports\cite{parmar2019and} are examples of HPSD, which are derived from Olympic competitions.

Classification in HPSD is important to attract people's attention and to highlight athlete's performance, and even to assist referees \cite{karpathy2014large}\cite{yue2015beyond}\cite{abu2016youtube}.
The significant difference between video classification and image classification is logical connection relationship in video clips \cite{bovik2010handbook}, which involve motion besides form (content) of frames.
Therefore, human action classification task should be implemented based on temporal human information without background rather than static action frame information.
H.kuehne \emph{et al.} \cite{kuehne2011hmdb} mentioned that the recognition rate of static joint locations alone in UCF sports \cite{rodriguez2008action} dataset is above $98\%$ while the use of joint kinematics is not necessary.
The result above obviously violates to the principle of video classification task.
Thus, both motion and pose are indispensable modalities in human action classification \cite{el2018human}.

Besides classification of HPSD, AQA is an important but unique task in HPSD.
Some AQA datasets have been proposed, such as MIT Olympic sports dataset\cite{pirsiavash2014assessing}, Nevada Olympic sports dataset\cite{parmar2019and} and AQA-7\cite{parmar2019action}, which predict the action performance of diving, gymnastic vault, etc.
In this field, Parmar \emph{et al.} \cite{parmar2017learning} adopts C3D features combining regression with the help of SVR\cite{drucker1997support} (LSTM\cite{hochreiter1997long}), which gets satisfactory results.
The development direction of HPSD is clearly given by the above works while AQA still remains to be developed as the materials of these datasets is still inadequate.
It could be concluded that the functions of datasets are gradually expanded with the development of video techniques \cite{kay2017kinetics}\cite{gu2018ava}\cite{perazzi2016benchmark}\cite{xu2016msr}\cite{sigal2006humaneva}.
However, it lacks temporal segmentation related datasets in recent years.
Our proposed FSD-10 is a response to these issues.
As far as we know, FSD-10 is the first dataset which gather actions task types including action classification, AQA and temporal segmentation.

\begin{figure*}[htbp]
	\centerline{\includegraphics[width=7in]{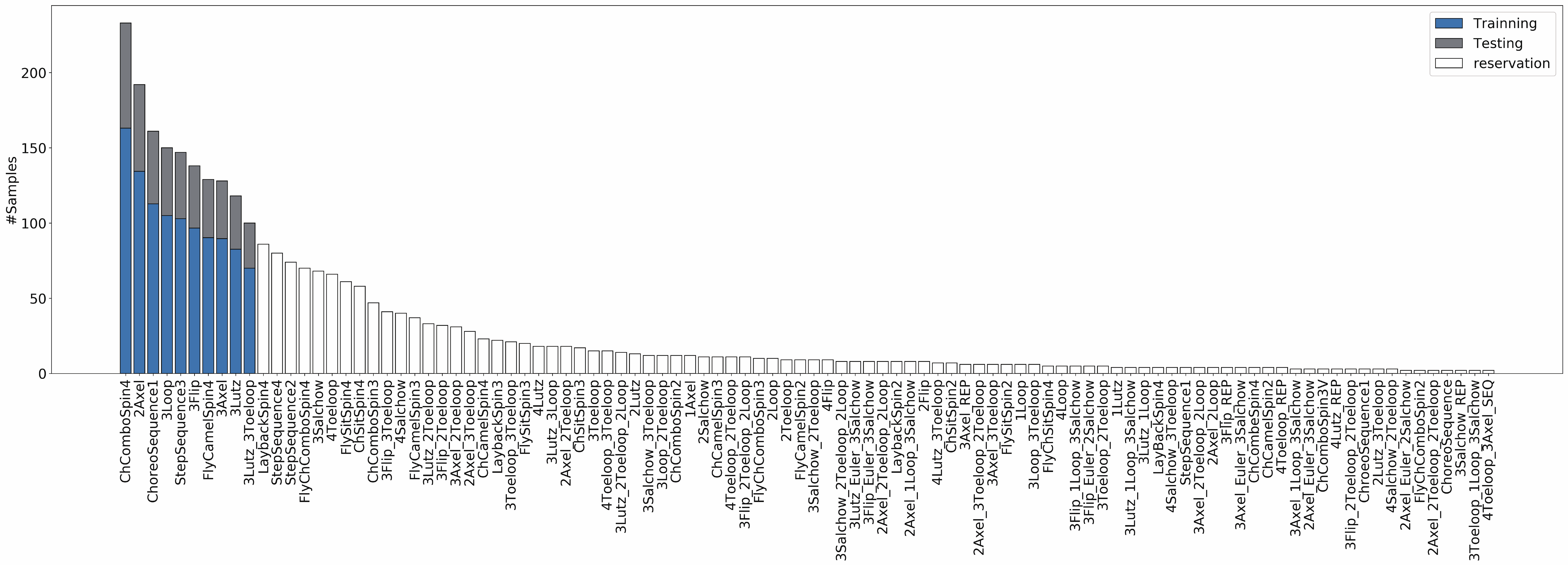}}
	\caption{Number of clips per action class. The horizontal axis indicates action types while the vertical axis indicates action amout.}
	\label{fig_hist}
\end{figure*}

\section{Figure Skating Dataset}

In this section, we describe details regarding the setup and protocol followed to capture our dataset.
Then, we discuss the temporal segmentation and assessment tasks of FSD-10 and its future extensions.

\subsection{Video Capture}

In FSD-10, all the videos are captured from the World Figure Skating Championships in 2017-2018, which originates from high-definition record on YouTube.
The source video materials have large ranges of fps (frame per second) and image size, which are standardized to 30 fps and 1080 $\times$ 720 to ensure the consistency of the dataset respectively.
As for the content, the programs of figure skating can be divided into four types, including men short program, ladies short program, men free skating and ladies free skating.
During the skating, camera is moving with the skater to ensure that the current action can be located center in videos.
Besides, auditorium and ice ground are collectively referred to the background, which are similar and consistent in all videos.

\subsection{Segmentation and Annotation}
\label{sec_segment}

\begin{table}[htbp]
	\caption{A single clip example of FSD-10}
	\setlength{\tabcolsep}{8mm}
	\label{table_clip}
\begin{tabular}{ccc}
	\toprule
	Action & Value \\
	\midrule
	Type & 2Axel\\
    BV & 3.3/3.63\\
    GOE & 0.58\\
    Skater name & Starr ANDREWS\\
    Skater gender & F\\
    Skater age & 16\\
    Coach & Derrick Delmore, Peter Kongkasem\\
    Music & Fever performed by Beyonce\\
	\bottomrule
\end{tabular}
\end{table}

To meet the demand of deep learning and other machine learning models, video materials are manually separated into action clips by undergraduate students.
Each clip is ranging from 3s to 30s, the principe of the segmentation is to completely reserve action content of videos, in other words, warm-up and ending actions are also included.
Actions type, base value (BV),  grade of execution (GOE), skater name, skater gender, skater age, coach and music are recorded in each clip.
A detailed record example of one clip is shown in Tab. \ref{table_clip}.
Unfortunately, the distribution of actions is out of tune, as different kinds of actions are totally 197 and the number of each class of action instances is from 1 to 223.
To avoid insufficient training, our FSD-10 is composed of the top 10 actions (as shown in Fig. \ref{fig_display}, Fig. \ref{fig_hist}).
Generally speaking, these selected actions could fall into three categories: jump, spin and sequence.
And in details, \emph{\textcolor[rgb]{0, 1, 0}{ChComboSpin4, FlyCamelSpin4}, \textcolor[rgb]{0, 0, 1}{ChoreoSequence1, StepSequence3}, \textcolor[rgb]{1, 0, 0}{2Axel, 3Loop, 3Flip, 3Axel, 3Lutz and 3Lutz$\_$3Toeloop}} are included.

\subsection{Action Discriminant Analysis}
\label{confusion}

Compared with existing sports, figure skating is hardly distinguished in action types.
The actions in figure skating look like tweedledum and tweedledee, which are difficult to distinguish in classification task.
(1) It is difficult to judge the action type by human pose in a certain frame or a few frames. For instance, Loop jump and Lutz jump are similar in warm-up period, it is impossible to distinguish them beyond take-off process.
(2) The difference between some actions exists on a subtle pose in fixed frames. For example, Flip jump and Lutz jump are both picking-ice actions, which are same in take-off with left leg and picking ice with right toe. The only subtle difference of the two types of jump is that takes-off process of Flip jump is by inner edge while Lutz jump is by outer edge.
(3) The spin direction is uncertain. The rotational direction of most skaters is anticlockwise while that of others is opposite (e.g. Denise Biellmann).
These factors of the above issues make classification task of FSD-10 a challengeable problem.

\subsection{Temporal Segmentation and Assessment}
\label{tasks}

\begin{itemize}
\item \textbf{Action Quality Assessment (AQA)}. AQA would be emerged as an imperative and challengeable issue in figure skating, which is used to evaluate performance level of skaters in skating programs by BV score plus GOE score.
As shown in Sec. \ref{sec_segment}, BV and GOE are included in our dataset, which are mainly depend on action types and skater's action performance respectively.
The key points of AQA are summarised as follows.
(1) BV is depend on action types and degree of action difficulty. Besides, $10\%$ bonus BV score is appended on the second half of a program.
(2) During the aerial process, insufficient number of turns and raising hands cause GOE deduction and bonus respectively.
(3) It is caused GOE deduction of one action by hand support, turn over, paralleling feet and trip during the landing process.

\item \textbf{Action Temporal Segmentation(ATS)}. It is indicated that ATS is also a significant issue of FSD-10.
A standard action is of certain stages.
For example during the process of a jump action, four stages are commonly included, which are preparation, take-off process, spinning in the air and landing process.
Separating these steps is a key technical issue for AQA and further related works.

\item \textbf{Long Sequence Segmentation and Assessment (LSSA)}. Another important issue is LSSA, which is different from AAS in sensitivity of actions order.
Action order and relation need to be considered in LSSA, which is summarised as follows.
(1) LSSA is the total of all the single action score, which is calculated similar to AQA for each action.
(2) It is caused a fail case \footnote{The twice appearance action  will have a zero score whether it is successful or not.} when the twice appearance of same action.
\end{itemize}

\subsection{Future Extension}

Our proposed FSD-10 is sensitive to domain knowledge and abundant in practical tasks.
Besides the tasks in Sec. \ref{tasks} , FSD-10 is convenient to be extended into other fields.
For example, pair skating is also a worth noting program in figure skating.
In the pair skating, to throw one skater by another is a classical item, and the action in the air could be illustrated as a corresponding jump action in men/ladies programs.
To transfer single skating models to pair skating ones is an interesting open problem, which could be deemed as a kind of transfer learning\cite{oquab2014learning}\cite{yosinski2014transferable} or meta learning\cite{lee2019meta}\cite{hsu2018unsupervised}.
Besides, action reasoning\cite{ginsberg1988reasoning}\cite{tuck2017theory} is an interesting issue.
For example, it is straightforward to conclude a 3Lutz-3Toeloop jump if single 3Lutz jump and 3toeloop jump have been recognized.
In addition, coordination between the action and the other sports elements such as program content assessment by rhythmical actions in time to music is also a focused issue.
Fortunately, the corresponding music of actions is included in FSD-10.
It is to be explored whether the music ups and downs is related to the action rhythm, which is an exciting cross-data-modality learning task.

\section{Keyframe based Temporal Segment Network (KTSN)}

In this section, we give a detailed description of our keyframe based temporal segment network.
Specifically, we first discuss the motivation of key frame sampling in Sec. \ref{motivation_kfs}.
Then, sampling method of key frame is proposed in Sec. \ref{HPS} and \ref{key_frame}.
Finally, network structure of KTSN is detailedly introduced in Sec. \ref{framework}.

\subsection{Motivation of Key Frame Sampling}
\label{motivation_kfs}

To solve the problem of long-range temporal modeling, a sampling method called segment based sampling is proposed by temporal segment networks(TSN)\cite{wang2018temporal}.
The main idea of segment based sampling is ``the dense recoded videos are unnecessary for content which changes slowly".
The method is essentially a kind of sparse and global sampling, which is significant in improving recognition performance of long-range videos.
But an intractable problem is caused that it is unbefitting for fast changing videos.
Especially for tasks like figure skating, which is complex in content and fast in speed.
It is illustrated in Sec. \ref{confusion} that actions of figure skating are similar in most of frames, and the only difference lies on certain frames.
Therefore, for sampling problem, we propose a method \emph{to sample key frames by domain knowledge and to sample other frames by segment based sampling}.
It is then focus on how to sample key frames in segmentation based sampling.

The problem motivates us to explore an indicator of sampling by domain knowledge in figure skating.
For example, by observing the process of jump, we find that arms and legs move following specific rules during the process.
Thus, we are encouraged to define an indicator using anatomical keypoints to measure this kind of rules, which is called \emph{Human Pose Scatter} and is illustrated in the next section.

\subsection{Human Pose Scatter (HPS)}
\label{HPS}

\begin{figure}[htbp]
	\centerline{\includegraphics[width=3in]{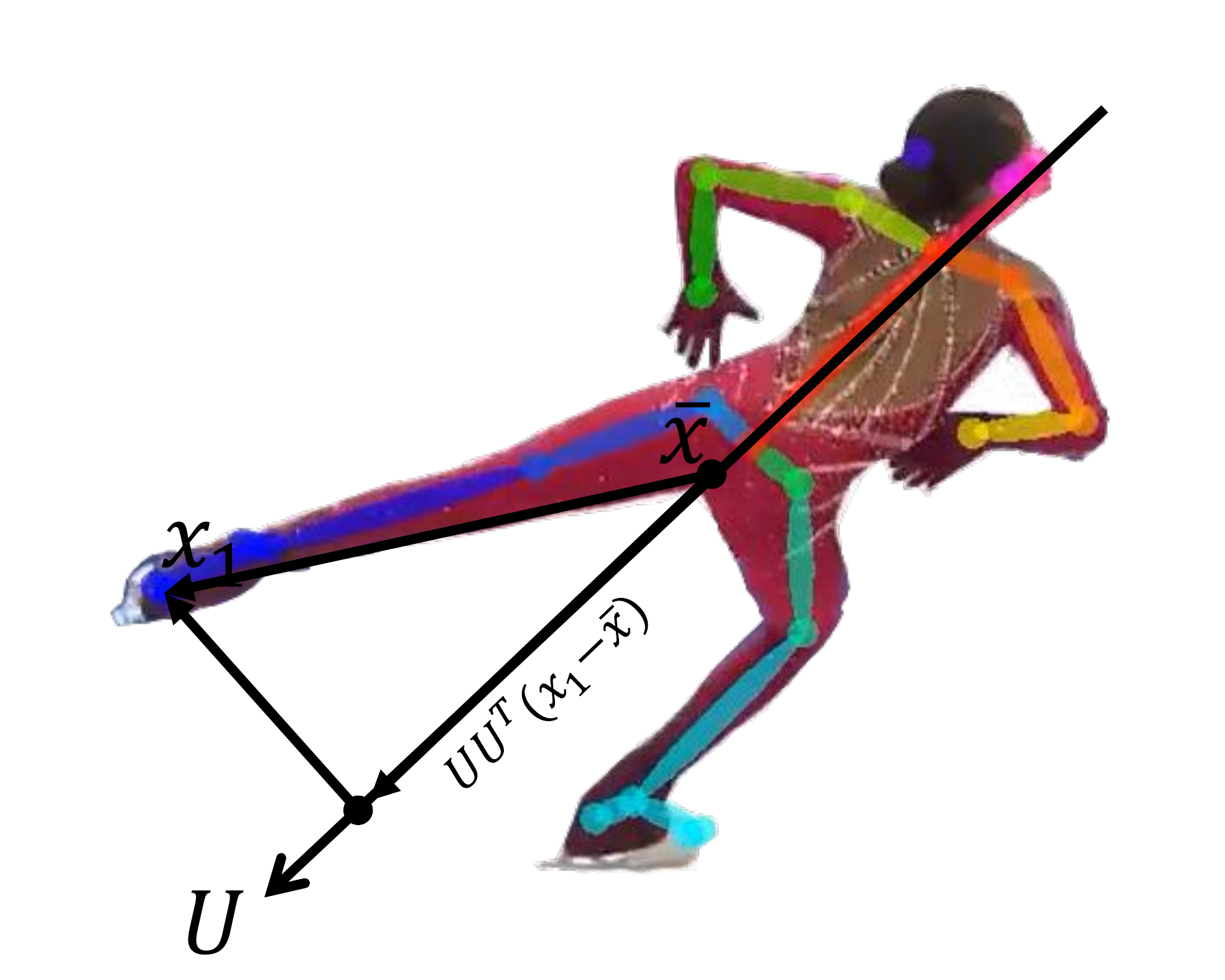}}
	\caption{The sketch map of human pose scatter. A 18-anatomical-keypoints recognition algorithm called open-pose\cite{cao2017realtime}\cite{cao2018openpose}\cite{simon2017hand}\cite{wei2016cpm} is adopted to get the spatial position and anatomical structure of human in one frame.}
	\label{fig_hps}
\end{figure}

Before key frame sampling, we will introduce HPS for a preparation.
HPS is an indicator which is denoted by the scatter of arms and legs relative to the body.
The main positions of human (left eye, right eye, etc. ) are marked by the open-pose algorithm, which is shown in Fig. \ref{fig_hps}.
Next, HPS is introduced in the rest of this section.

As shown in Fig. \ref{fig_hps}, given a frame matrix of the responding anatomical structure, $X = [x_1, \cdots, x_n] \in R^{D \times n}$ (in this paper $D = 2$ and $n = 18$), in which each row corresponds to a particular feature while each column corresponds to a human anatomical keypoint, and $\bar{x}$ is the column mean value of $X$.
Then, $\hat{X}$ can be defined as $\hat{X} = [x_1 - \bar{x}, x_2 - \bar{x}, \cdots, x_i - \bar{x}]$.
Motivated by the principal component analysis (PCA\cite{wold1987principal}), HPS can be defined by Eq. \ref{eq_1}.
Projection matrix $U$ ($U_{D \times d} = [u_1, u_2, \cdots, u_d]$, satisfying $U^{T}U = I$, which leads $I-UU^T$ the idemfactor) can be calculated by eigenvalue decomposition (EVD) of $\hat{X}\hat{X}^T$.

\begin{equation}
\label{eq_1}
\zeta_{HPS} = \sum_{i=1}^{n}\left \| (x_i - \bar x) - UU^T(x_i - \bar x) \right \|^{2}_{2}
\end{equation}

Actually, Eq. \ref{eq_1}, only a formal expression of $\zeta_{HPS}$, can be easily rewritten as Eq. \ref{eq_2} to simplify calculation.
In Eq. \ref{eq_2}, $Tr\begin{bmatrix} \hat{X}\hat{X}^T \end{bmatrix}$ and $Tr\begin{bmatrix} U^T\hat{X}\hat{X}^TU \end{bmatrix} = \sum_{j=1}^{d}\lambda_{j}$\footnote{$\lambda_{j}$ is the $j$-th eigenvalue of covariance matrix $\hat{X}\hat{X}^T$, in our paper $d=1$.} have been calculated during the EVD process.
Therefore, Eq. \ref{eq_2} offers an efficient approach for calculation of $\zeta_{HPS}$.

\begin{equation}
\label{eq_2}
\begin{split}
\zeta_{HPS} = &\sum_{i=1}^{n}\left \| (I - UU^T)(x_i - \bar{x})\right \|_2^2 \\
= &\sum_{i=1}^{n}(x_i - \bar{x})^T(I-UU^T)(x_i-\bar{x}) \\
= & Tr(\hat{X}\hat{X}^T) - \sum_{j=1}^{d}\lambda_{j}
\end{split}
\end{equation}

$\zeta_{HPS}$ is the sum of variance towards the principal axis of human anatomical keypoints, which is used to reflect the degree of human pose stretching.
Therefore, $\zeta_{HPS}$ is a tremendous help in finding key frames of sequences.
For example, during a process of jump, $\zeta_{HPS}$ of spin in the air process is lower than it of the take off process, and extreme point maps a turning point responding to some certain actions.
This can be referred to Fig. \ref{fig_first}.

\subsection{Key Frame Sampling}
\label{key_frame}

\begin{figure}[htbp]
	\centerline{\includegraphics[width=3.4in]{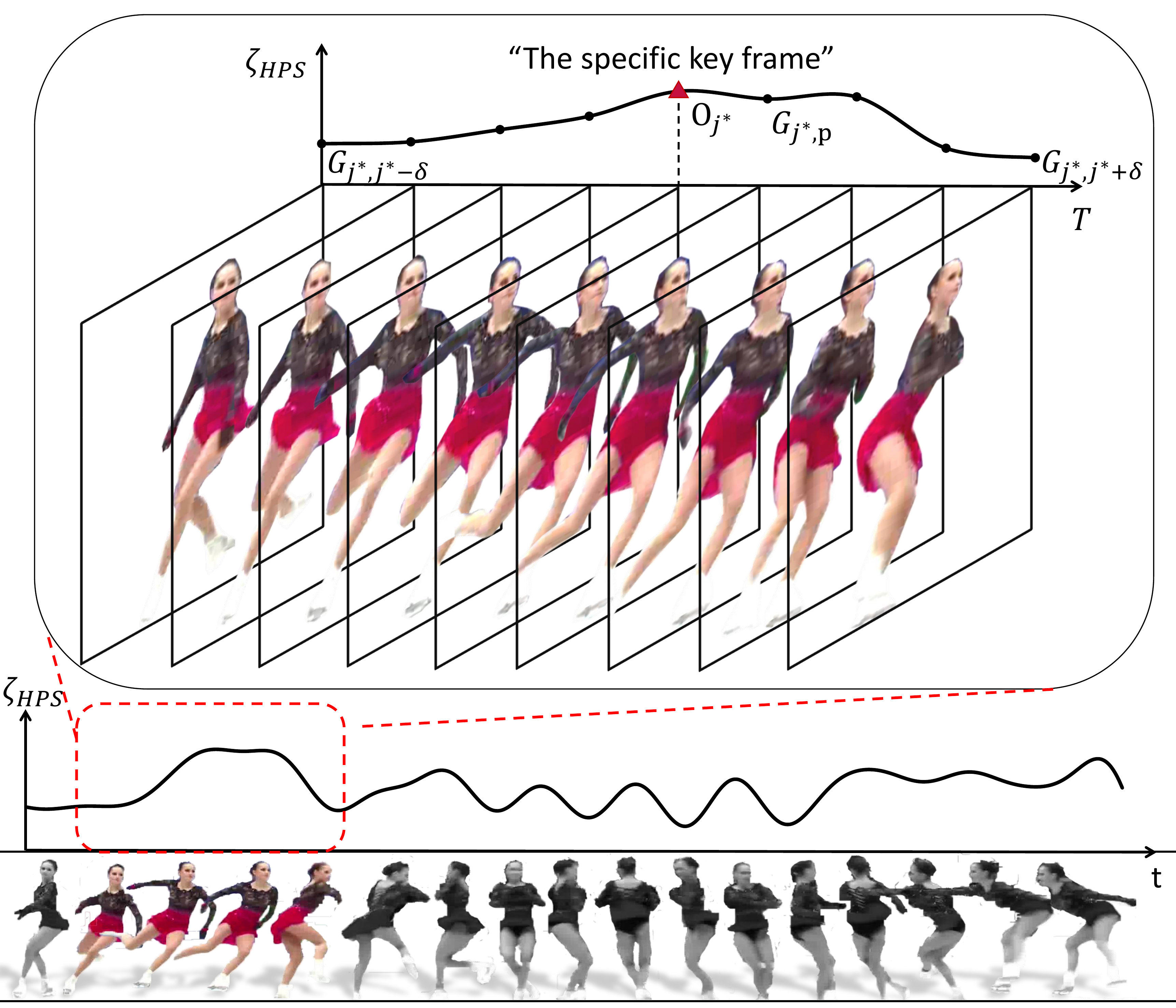}}
	\caption{The sketch map of key frame sampling. A 2-Axel jump is performed by Alina Ilnazovna Zagitova, which is a same action with Fig. \ref{fig_first}. $t$ represents the time variable.}
	\label{fig_sample}
\end{figure}

\begin{figure*}[htbp]
	\centerline{\includegraphics[width=7in]{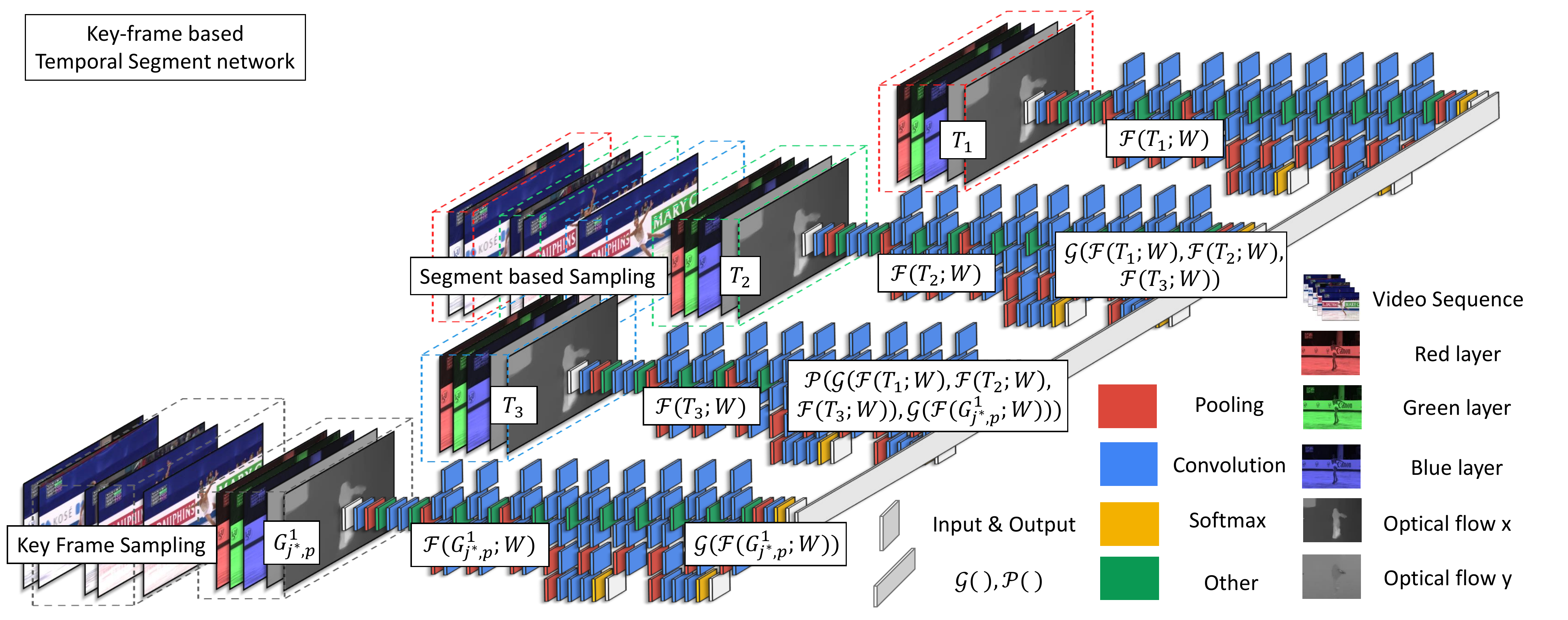}}
	\caption{Keyframe based temporal segment network. One input video is divided into $k$ segments (here we show the $k$ = 3 case) and a short snippet is randomly selected from each segment. Meanwhile, $n$ key frames are selected from the Video (here $L$ = 1). The snippets are represented by modalities such as RGB frames and Optical flow. Then these snippets are fused by an aggregation function $\mathcal{G}$. Last, $\mathcal{P}$ combines the outputs and makes video-level predictions. ConvNets on all snippets share parameters.}
	\label{fig_KTSN}
\end{figure*}

Key frame sampling is an important issue in video analysis of figure skating, and the key frames should contain most discriminant information of one video.
In the figure skating task, the fast changing motion frames are distinctly important for jump actions.
As illustrated in Sec. \ref{HPS}, the kind of fast changing motion could be quantified as $\zeta_{HPS}$ by using anatomical keypoints.
Therefore, in a $T$ frames clip of a video, the frame corresponding to extreme frame $O_{j^*}$, which satisfies $j^*=\argmax\limits_{j} \zeta_{HPS}^j$, is defined as ``the specific key frame", where $\delta$ indicates the near frame radius of $O_{j^*}$, $ j \in \left \{\delta+1, \cdots, T-\delta \right \}$.
In addition, as nearby $\delta$ frames of $O_{j^*}$, $G_{j^*}= \left \{{G_{j^*, j^*-\delta}}, \cdots, G_{j^*, p}, \cdots, G_{j^*, j^* + \delta} \right \}$ are defined as ``key frames", where $p \in \left \{ {j^* - \delta, \cdots, j^* + \delta} \right \} \backslash j^*$(see Fig. \ref{fig_sample}).
In a multi-action clip,  we will get $O^l_{j^*}$, where $l=1, 2, \cdots, L$, $L$ indicates the number of extreme frames.
Then, key frame sampling could be described as follows.
First, finding the specific key frame(s) in action sequences.
Second, sampling the video from the specific key frame(s) and its nearby frames using the similar approach of TSN.
In one batch, TSN samples a random frame while key frame sampling involves a random frame $G_{j*, p}^l$ in $G_{j^*} \cup O_{j^*}$.

\subsection{Framework Structure}
\label{framework}

An effective and efficient video-level framework, coined keyframe based Temporal Segment Network (KTSN, Fig. \ref{fig_KTSN}), is designed with the help of key frame sampling.
Compared with single frame or short frame pieces, KTSN works on short snippets sampled from the videos.
To make these short snippets representative for the entire videos as well as to keep computational consumption steerable, the video is divided into several segments first.
Then, sampling is continued on each piece of snippets.
As illustrated in \ref{motivation_kfs}, our sampling methods are divided to two parts, which are segment based sampling\cite{wang2018temporal} and key frame sampling respectively.
With the sampling method, each piece contributes its snippet-level predictions of actions classes.
And these snippet-level predictions are aggregated as video-level scores by a consensus function.
The video-level prediction is more accurate than any single snippet-level predictions for the reason that it captures details of the whole videos.
During the process of training, the model is iteratively updated in video-level as follows.

We divide a video into $\emph{k}$ segments ($V_{video} = \left \{ \alpha_1,\alpha_2, \cdots, \alpha_k \right \}$ of equal duration pieces.
One frame $T_k$ is randomly selected from its corresponding segment $\alpha_k$.
Similarly, One snippet $G_{j^*, p}^l$ is randomly selected from its corresponding $G^l_{j^*} \cup O^l_{j^*}$, which have been proposed in section \ref{HPS} and \ref{key_frame}.
The sequence of snippets ($T_1, T_2, \cdots, T_k$ and $G_{j^*, p}^1, G_{j^*, p}^2, \cdots, G_{j^*, p}^L$) makes up of the model of keyframe based temporal segment network as follows.

\begin{equation}
\label{eq_3}
\begin{split}
KTSN(T_1, &T_2, \cdots, T_k, G_{j^*, p}^1, G_{j*, p}^2, \cdots, G_{j^*, p}^L)= \\
&\mathcal{P}(\mathcal{G}(\mathcal{F}(T_1; W), \mathcal{F}(T_2; W), \cdots, \mathcal{F}(T_k; W)), \\
\mathcal{G}(\mathcal{F}&(G_{j^*, p}^1; W), \mathcal{F}(G_{j^*, p}^2; W), \cdots, \mathcal{F}(G_{j^*, p}^L; W)))
\end{split}
\end{equation}

In Eq. \ref{eq_3}, $\mathcal{F}(T_k; W)$ (or $\mathcal{F}(G_{j*, p}^l; W)$) is the function on behalf of a ConvNet\cite{lo1995artificial} with model parameters $\emph{W}$ which plays an important role in the snippet $T_k$ ($G_{j*, p}^l$) and obtain a consensus of classification hypothesis.
It is given the prediction of the whole video by an aggregation function $\mathcal{G}$, which combines all the output of individual snippets.
Based on the value of $\mathcal{G}$, $\mathcal{P}$ works out the classification results.
In this paper, $\mathcal{P}$ is represented as softmax function\cite{memisevic2010gated}.

\section{Experiments}

In order to provide a benchmark for our FSD-10 dataset, we evaluate various approaches under three different modalities: RGB, optical flow and anatomical keypoints (skeleton).
We also conduct experiments on cross dataset validation.
The following describes the details of our experiments and results.

\subsection{Evaluations of Methods on FSD-10}

\begin{table}[htbp]
	\caption{Performance of baseline classification  models evaluates with FSD-10. The short names of these methods are two-stream inflated 3D ConvNet\cite{carreira2017quo} (Inception-v1\cite{szegedy2015going}, Inception-v3\cite{szegedy2016rethinking}, ResNet-50\cite{he2016deep} and ResNet-101\cite{he2016deep}), spatial temporal graph convolution networks (st-gcn\cite{yan2018spatial}), densenet with anatomical keypoints networks\cite{huang2017densely}, temporal segment network\cite{wang2018temporal} and keyframe based temporal segment network respectively. DenseNet+ denotes DenseNet with centralization. Besides, TSN(7) represents TSN($k$=7) and KTSN(6,1) represents KTSN($k$=6, $L$=1) corresponding to Sec. \ref{framework}, similarly hereinafter.}
	\setlength{\tabcolsep}{3mm}
	\label{table_exp}
\begin{center}
\begin{tabular}{cccccc}
	\toprule
    & RGB & Optical flow & Skeleton & Fusion \\
    \midrule
    Inception-v1\cite{szegedy2015going} & 43.765 & 52.706  & - & 55.052 \\
    Inception-v3\cite{szegedy2016rethinking} & 50.118 & 57.245  & - & 59.624 \\
    Resnet-50\cite{he2016deep} & 62.550 & 58.824  & - & 64.941 \\
    Resnet-101\cite{he2016deep} & 78.823 & - & 61.549 & 79.851 \\
    St-gcn\cite{yan2018spatial} & - & -  & 72.429 &  - \\
    DenseNet\cite{huang2017densely} & - & - & 74.353 &  - \\
    DenseNet+\cite{huang2017densely} & - & - & 79.765 & - \\
    \midrule
    TSN(7)\cite{wang2018temporal} & 53.176 & 75.165 & - & 76.000 \\
    KTSN(6, 1) & 56.941 & 76.941 & - & 77.412 \\
    \midrule
    TSN(9)\cite{wang2018temporal} & 59.294 & 80.235 & - & 80.235 \\
    KTSN(8, 1) & 59.529 & 80.471 & - & 80.471 \\
    \midrule
    TSN(11)\cite{wang2018temporal} & 59.294 & 82.118 & - & 82.118 \\
    \textbf{KTSN(10, 1)} & \textbf{63.294} & \textbf{82.588} & - & \textbf{82.588} \\
	\bottomrule
\end{tabular}
\end{center}
\end{table}

The state-of-the-art classification methods of human action are evaluated on FSD-10 in this section.
In order to make the experiments representative in action recognition field, 2D methods and 3D methods are selected respectively.
TSN (or KTSN) is selected as a 2D method, which is a typical two stream structure.
As for 3D methods, two-stream inflated 3D ConvNet (I3D) with Inception-v1\cite{szegedy2015going}, Inception-v3\cite{szegedy2016rethinking} and Resnet-50, Resnet-101\cite{he2016deep} are involved for comparison.
Besides, anatomical keypoints related methods (st-gcn\cite{yan2018spatial}, densenet\cite{huang2017densely}) are tested on our dataset for reference.
As listed in Tab. \ref{table_exp}, our proposed KTSN achieves state-of-the-art results on FSD-10.
KTSN method is raised as a baseline for this high-speed dataset.

\subsection{Evaluations between TSN and KTSN}

The experimental results in Tab. \ref{table_exp} show that key frame sampling is effective in sparse sampling.
As illustrated above, key frame is crucial for competitive sports classification tasks.
Especially for jump actions, it is vital to choose take off frames, which enhances discriminant of different jump actions.
An example of key frame sampling of jump actions is shown in Fig. \ref{fig_sample}.
By catching these key frames instead of random sampling, KTSN achieves better classification performance than TSN\cite{wang2018temporal}.

\subsection{Cross-dataset Validations}

To verify motion characteristic of a sport video, the cross-dataset validation is conducted on both classical dataset UCF101 and FSD-10.
RGB and optical flow features are two typical ways for action recognition.
Generally speaking, the RGB feature based method places emphasis on human pose while the optical flow based method attaches importance to transformation between actions.
It is illustrated in Tab. \ref{table_cross} that the performance differences between RGB and Optical flow of FSD-10 is greater than UCF101, which indicates motion is more important than human static pose in FSD-10.
Besides, in sampling stage, FSD-10 is more sensitive to fine-grained segmentation, as the performance raises $5.4\%$ compared with $2.5\%$ in UCF101 (3-segment-sampling to 9-segment-sampling).
In FSD-10, actions are sensitive to its motion rather than its pose, which is consistent with the original intention of the dataset.

\begin{table}[htbp]
	\caption{Results of cross-dataset generalization. UCF101 and FSD-10 are trained with TSN($k$=3) and TSN($k$=9) respectively.}
	\setlength{\tabcolsep}{3mm}
	\label{table_cross}
\begin{center}
\begin{tabular}{ccccc}
	\toprule
    Dataset & Methods & RGB & Optical flow & Fusion \\
    \midrule
    \multirow{2}{*}{UCF101} & TSN(3) & 85.673 & 85.170 &  92.466 \\
     & TSN(9) & 86.217 & 89.713 & $\textbf{94.909}$\\
    \midrule
    \multirow{2}{*}{FSD-10} & TSN(3) & 48.941 & 75.765 & 76.765 \\
     & TSN(9) & 59.294 & 82.118 & $\textbf{82.118}$\\
	\bottomrule
\end{tabular}
\end{center}
\end{table}

\section{Conclusion}

In this paper, we build an action dataset for competitive sports analysis, which is characterised by high action speed and complex action content.
We find that motion is more valuable than form (content and background) in this task.
Therefore, compared with other related datasets, our dataset focuses on the action itself rather background.
Our dataset creates many interesting tasks, such as fine-grained action classification, action quality assessment and action temporal segmentation.
Moreover, the dataset could even be extended to pair skating in transfer learning and coordination between action and music in cross-data-modality learning.

We also present a baseline for action classification in FSD-10.
Note that domain knowledge is significant for competitive action classification task.
Motivated by this idea, we adopt key-frame based temporal segmentation sampling.
It is illustrated that action sampling should focus more on related frames of motion.
The experimental results demonstrate that the key frame is effective for skating tasks.

In the future, more diversified action types and tasks are intended to be included to our dataset, and more interesting action recognition methods are likely to be proposed with the help of FSD-10.
We believe that our dataset will promote the development of action analysis and related research topics.

%

\section*{Supplementary Material}

%

\subsection*{1. Training of KTSN}

\textbf{Architectures} Inception network, chosen in KTSN, is a milestone in CNN classifier.
The main characteristics of inception network is the inception module, which is designed to process the features in different dimensionality (such as the 1$\times$1, 3$\times$3 and 5$\times$5 convolution core) parallelly.
The result of Inception network is a better representation than single-layer-networks, which is excellent in local topological structure.
Specifically, Inception v3 is selected in the experiments as follows.

\textbf{Input} Our model is based on spatial and temporal level, which are corresponding to RGB frames\cite{susstrunk1999standard} and optical flow\cite{horn1981determining} respectively.
RGB mode is an industrial standard of color, combining \textit{R}ed, \textit{G}reen and \textit{B}lue into chromatic frames, which is also called the spatial flow.
Besides, optical flow mode derives from calculating the differences between adjacent frames, which indicates temporal flow.

\begin{figure}[htbp]
	\centerline{\includegraphics[width=3.2in]{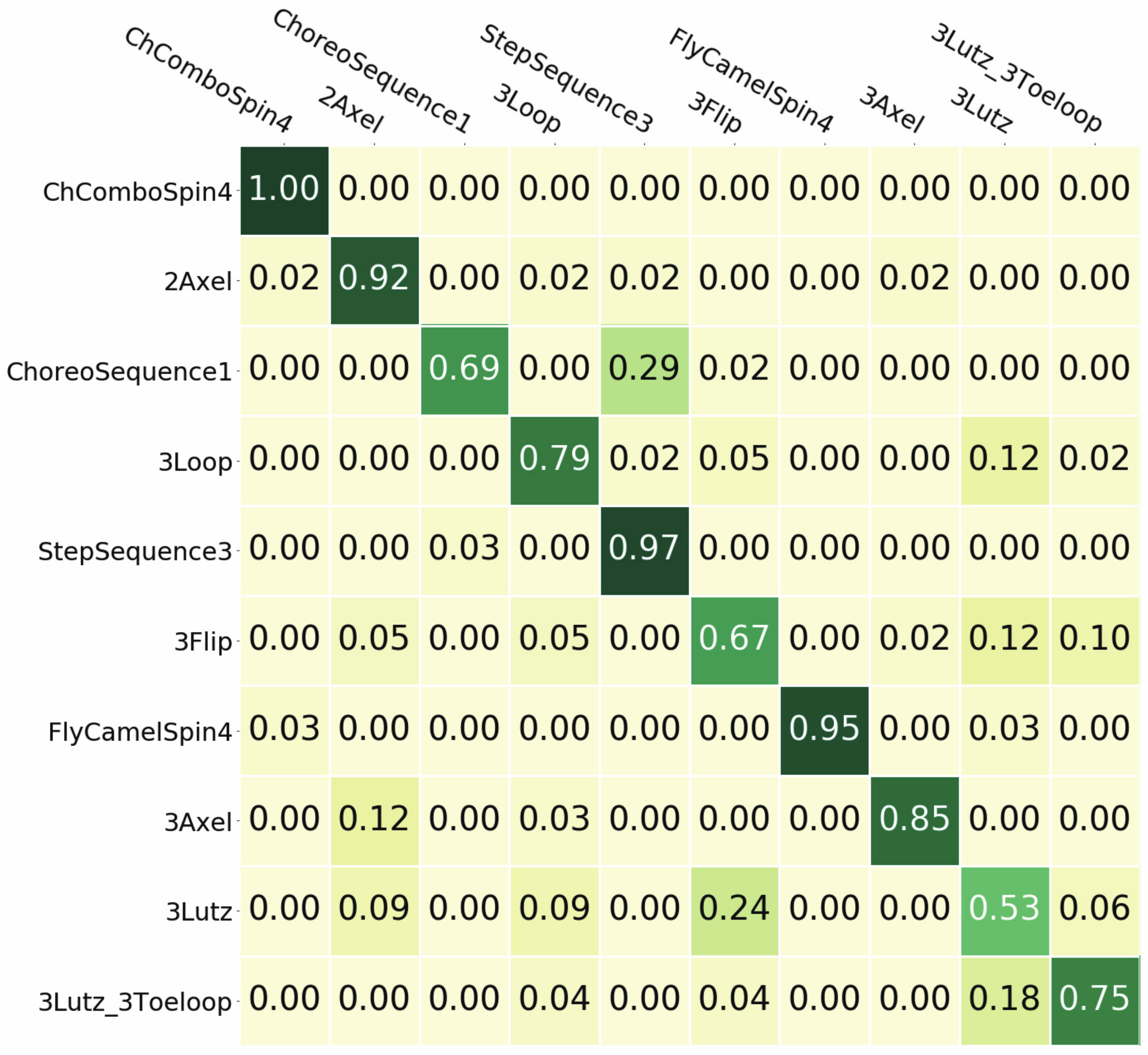}}
	\caption{The confusion matrix results of FSD-10}
	\label{fig_matrix}
\end{figure}

\subsection*{2. Confusion Matrix of Actions}

It's significant to involve domain knowledge in competitive sports content analysis.
It is illustrated the confusion relationships between actions by examples in Sec. 3.3, and this part is a validation, as shown in Fig. \ref{fig_matrix}.
First, the skating spin obtains the best classification results, this is because it could judge a skating spin of video precisely with a few isolated frames using deep networks.
Second, the skating jump is the most difficult to distinguish.
It's worth note that up to $24\%$ Lutz jump actions are identified as Flip jump actions.

{\small
\bibliographystyle{ieee_fullname}
\bibliography{egbib}

\begin{thebibliography}{10}\itemsep=-1pt

\bibitem{abu2016youtube}
Sami Abu-El-Haija, Nisarg Kothari, Joonseok Lee, Paul Natsev, George Toderici,
  Balakrishnan Varadarajan, and Sudheendra Vijayanarasimhan.
\newblock Youtube-8m: A large-scale video classification benchmark.
\newblock {\em arXiv preprint arXiv:1609.08675}, 2016.

\bibitem{bovik2010handbook}
Alan~C Bovik.
\newblock {\em Handbook of image and video processing}.
\newblock Academic press, 2010.

\bibitem{cao2018openpose}
Zhe Cao, Gines Hidalgo, Tomas Simon, Shih-En Wei, and Yaser Sheikh.
\newblock Open{P}ose: realtime multi-person 2{D} pose estimation using {P}art
  {A}ffinity {F}ields.
\newblock In {\em arXiv preprint arXiv:1812.08008}, 2018.

\bibitem{cao2017realtime}
Zhe Cao, Tomas Simon, Shih-En Wei, and Yaser Sheikh.
\newblock Realtime multi-person 2d pose estimation using part affinity fields.
\newblock In {\em CVPR}, 2017.

\bibitem{carreira2017quo}
Joao Carreira and Andrew Zisserman.
\newblock Quo vadis, action recognition? a new model and the kinetics dataset.
\newblock In {\em proceedings of the IEEE Conference on Computer Vision and
  Pattern Recognition}, pages 6299--6308, 2017.

\bibitem{drucker1997support}
Harris Drucker, Christopher~JC Burges, Linda Kaufman, Alex~J Smola, and
  Vladimir Vapnik.
\newblock Support vector regression machines.
\newblock In {\em Advances in neural information processing systems}, pages
  155--161, 1997.

\bibitem{el2018human}
Hany El-Ghaish, Mohamed~E Hussein, Amin Shoukry, and Rikio Onai.
\newblock Human action recognition based on integrating body pose, part shape,
  and motion.
\newblock {\em IEEE Access}, 6:49040--49055, 2018.

\bibitem{ginsberg1988reasoning}
Matthew~L Ginsberg and David~E Smith.
\newblock Reasoning about action i: A possible worlds approach.
\newblock {\em Artificial intelligence}, 35(2):165--195, 1988.

\bibitem{gu2018ava}
Chunhui Gu, Chen Sun, David~A Ross, Carl Vondrick, Caroline Pantofaru, Yeqing
  Li, Sudheendra Vijayanarasimhan, George Toderici, Susanna Ricco, Rahul
  Sukthankar, et~al.
\newblock Ava: A video dataset of spatio-temporally localized atomic visual
  actions.
\newblock In {\em Proceedings of the IEEE Conference on Computer Vision and
  Pattern Recognition}, pages 6047--6056, 2018.

\bibitem{gudmundsson2017spatio}
Joachim Gudmundsson and Michael Horton.
\newblock Spatio-temporal analysis of team sports.
\newblock {\em ACM Computing Surveys (CSUR)}, 50(2):22, 2017.

\bibitem{he2016deep}
Kaiming He, Xiangyu Zhang, Shaoqing Ren, and Jian Sun.
\newblock Deep residual learning for image recognition.
\newblock In {\em Proceedings of the IEEE conference on computer vision and
  pattern recognition}, pages 770--778, 2016.

\bibitem{hochreiter1997long}
Sepp Hochreiter and J{\"u}rgen Schmidhuber.
\newblock Long short-term memory.
\newblock {\em Neural computation}, 9(8):1735--1780, 1997.

\bibitem{horn1981determining}
Berthold~KP Horn and Brian~G Schunck.
\newblock Determining optical flow.
\newblock {\em Artificial intelligence}, 17(1-3):185--203, 1981.

\bibitem{hsu2018unsupervised}
Kyle Hsu, Sergey Levine, and Chelsea Finn.
\newblock Unsupervised learning via meta-learning.
\newblock {\em arXiv preprint arXiv:1810.02334}, 2018.

\bibitem{huang2017densely}
Gao Huang, Zhuang Liu, Laurens Van Der~Maaten, and Kilian~Q Weinberger.
\newblock Densely connected convolutional networks.
\newblock In {\em Proceedings of the IEEE conference on computer vision and
  pattern recognition}, pages 4700--4708, 2017.

\bibitem{karpathy2014large}
Andrej Karpathy, George Toderici, Sanketh Shetty, Thomas Leung, Rahul
  Sukthankar, and Li Fei-Fei.
\newblock Large-scale video classification with convolutional neural networks.
\newblock In {\em Proceedings of the IEEE conference on Computer Vision and
  Pattern Recognition}, pages 1725--1732, 2014.

\bibitem{kay2017kinetics}
Will Kay, Joao Carreira, Karen Simonyan, Brian Zhang, Chloe Hillier, Sudheendra
  Vijayanarasimhan, Fabio Viola, Tim Green, Trevor Back, Paul Natsev, et~al.
\newblock The kinetics human action video dataset.
\newblock {\em arXiv preprint arXiv:1705.06950}, 2017.

\bibitem{kuehne2011hmdb}
Hildegard Kuehne, Hueihan Jhuang, Est{\'\i}baliz Garrote, Tomaso Poggio, and
  Thomas Serre.
\newblock Hmdb: a large video database for human motion recognition.
\newblock In {\em 2011 International Conference on Computer Vision}, pages
  2556--2563. IEEE, 2011.

\bibitem{lee2019meta}
Kwonjoon Lee, Subhransu Maji, Avinash Ravichandran, and Stefano Soatto.
\newblock Meta-learning with differentiable convex optimization.
\newblock In {\em Proceedings of the IEEE Conference on Computer Vision and
  Pattern Recognition}, pages 10657--10665, 2019.

\bibitem{lo1995artificial}
Shih-Chung~B Lo, Heang-Ping Chan, Jyh-Shyan Lin, Huai Li, Matthew~T Freedman,
  and Seong~K Mun.
\newblock Artificial convolution neural network for medical image pattern
  recognition.
\newblock {\em Neural networks}, 8(7-8):1201--1214, 1995.

\bibitem{marszalek2009actions}
Marcin Marsza{\l}ek, Ivan Laptev, and Cordelia Schmid.
\newblock Actions in context.
\newblock In {\em CVPR 2009-IEEE Conference on Computer Vision \& Pattern
  Recognition}, pages 2929--2936. IEEE Computer Society, 2009.

\bibitem{memisevic2010gated}
Roland Memisevic, Christopher Zach, Marc Pollefeys, and Geoffrey~E Hinton.
\newblock Gated softmax classification.
\newblock In {\em Advances in neural information processing systems}, pages
  1603--1611, 2010.

\bibitem{oquab2014learning}
Maxime Oquab, Leon Bottou, Ivan Laptev, and Josef Sivic.
\newblock Learning and transferring mid-level image representations using
  convolutional neural networks.
\newblock In {\em Proceedings of the IEEE conference on computer vision and
  pattern recognition}, pages 1717--1724, 2014.

\bibitem{parmar2019action}
Paritosh Parmar and Brendan Morris.
\newblock Action quality assessment across multiple actions.
\newblock In {\em 2019 IEEE Winter Conference on Applications of Computer
  Vision (WACV)}, pages 1468--1476. IEEE, 2019.

\bibitem{parmar2019and}
Paritosh Parmar and Brendan~Tran Morris.
\newblock What and how well you performed? a multitask learning approach to
  action quality assessment.
\newblock In {\em Proceedings of the IEEE Conference on Computer Vision and
  Pattern Recognition}, pages 304--313, 2019.

\bibitem{parmar2017learning}
Paritosh Parmar and Brendan Tran~Morris.
\newblock Learning to score olympic events.
\newblock In {\em proceedings of the IEEE Conference on Computer Vision and
  Pattern Recognition Workshops}, pages 20--28, 2017.

\bibitem{perazzi2016benchmark}
Federico Perazzi, Jordi Pont-Tuset, Brian McWilliams, Luc Van~Gool, Markus
  Gross, and Alexander Sorkine-Hornung.
\newblock A benchmark dataset and evaluation methodology for video object
  segmentation.
\newblock In {\em Proceedings of the IEEE Conference on Computer Vision and
  Pattern Recognition}, pages 724--732, 2016.

\bibitem{pirsiavash2014assessing}
Hamed Pirsiavash, Carl Vondrick, and Antonio Torralba.
\newblock Assessing the quality of actions.
\newblock In {\em European Conference on Computer Vision}, pages 556--571.
  Springer, 2014.

\bibitem{rodriguez2008action}
Mikel~D Rodriguez, Javed Ahmed, and Mubarak Shah.
\newblock Action mach a spatio-temporal maximum average correlation height
  filter for action recognition.
\newblock In {\em CVPR}, volume~1, page~6, 2008.

\bibitem{safdarnejad2015sports}
Seyed~Morteza Safdarnejad, Xiaoming Liu, Lalita Udpa, Brooks Andrus, John Wood,
  and Dean Craven.
\newblock Sports videos in the wild (svw): A video dataset for sports analysis.
\newblock In {\em 2015 11th IEEE International Conference and Workshops on
  Automatic Face and Gesture Recognition (FG)}, volume~1, pages 1--7. IEEE,
  2015.

\bibitem{shih2017survey}
Huang-Chia Shih.
\newblock A survey of content-aware video analysis for sports.
\newblock {\em IEEE Transactions on Circuits and Systems for Video Technology},
  28(5):1212--1231, 2017.

\bibitem{sigal2006humaneva}
Leonid Sigal and Michael~J Black.
\newblock Humaneva: Synchronized video and motion capture dataset for
  evaluation of articulated human motion.
\newblock {\em Brown Univertsity TR}, 120, 2006.

\bibitem{simon2017hand}
Tomas Simon, Hanbyul Joo, Iain Matthews, and Yaser Sheikh.
\newblock Hand keypoint detection in single images using multiview
  bootstrapping.
\newblock In {\em CVPR}, 2017.

\bibitem{soomro2012ucf101}
Khurram Soomro, Amir~Roshan Zamir, and Mubarak Shah.
\newblock Ucf101: A dataset of 101 human actions classes from videos in the
  wild.
\newblock {\em arXiv preprint arXiv:1212.0402}, 2012.

\bibitem{susstrunk1999standard}
Sabine S{\"u}sstrunk, Robert Buckley, and Steve Swen.
\newblock Standard rgb color spaces.
\newblock In {\em Color and Imaging Conference}, volume 1999, pages 127--134.
  Society for Imaging Science and Technology, 1999.

\bibitem{szegedy2015going}
Christian Szegedy, Wei Liu, Yangqing Jia, Pierre Sermanet, Scott Reed, Dragomir
  Anguelov, Dumitru Erhan, Vincent Vanhoucke, and Andrew Rabinovich.
\newblock Going deeper with convolutions.
\newblock In {\em Proceedings of the IEEE conference on computer vision and
  pattern recognition}, pages 1--9, 2015.

\bibitem{szegedy2016rethinking}
Christian Szegedy, Vincent Vanhoucke, Sergey Ioffe, Jon Shlens, and Zbigniew
  Wojna.
\newblock Rethinking the inception architecture for computer vision.
\newblock In {\em Proceedings of the IEEE conference on computer vision and
  pattern recognition}, pages 2818--2826, 2016.

\bibitem{tuck2017theory}
Mary Tuck and David Riley.
\newblock The theory of reasoned action: A decision theory of crime.
\newblock In {\em The reasoning criminal}, pages 156--169. Routledge, 2017.

\bibitem{van2010race}
Jacco Van~Sterkenburg, Annelies Knoppers, and Sonja De~Leeuw.
\newblock Race, ethnicity, and content analysis of the sports media: A critical
  reflection.
\newblock {\em Media, Culture \& Society}, 32(5):819--839, 2010.

\bibitem{wang2018temporal}
Limin Wang, Yuanjun Xiong, Zhe Wang, Yu Qiao, Dahua Lin, Xiaoou Tang, and Luc
  Van~Gool.
\newblock Temporal segment networks for action recognition in videos.
\newblock {\em IEEE transactions on pattern analysis and machine intelligence},
  2018.

\bibitem{wei2016cpm}
Shih-En Wei, Varun Ramakrishna, Takeo Kanade, and Yaser Sheikh.
\newblock Convolutional pose machines.
\newblock In {\em CVPR}, 2016.

\bibitem{wold1987principal}
Svante Wold, Kim Esbensen, and Paul Geladi.
\newblock Principal component analysis.
\newblock {\em Chemometrics and intelligent laboratory systems}, 2(1-3):37--52,
  1987.

\bibitem{xu2016msr}
Jun Xu, Tao Mei, Ting Yao, and Yong Rui.
\newblock Msr-vtt: A large video description dataset for bridging video and
  language.
\newblock In {\em Proceedings of the IEEE conference on computer vision and
  pattern recognition}, pages 5288--5296, 2016.

\bibitem{yan2018spatial}
Sijie Yan, Yuanjun Xiong, and Dahua Lin.
\newblock Spatial temporal graph convolutional networks for skeleton-based
  action recognition.
\newblock In {\em Thirty-Second AAAI Conference on Artificial Intelligence},
  2018.

\bibitem{yosinski2014transferable}
Jason Yosinski, Jeff Clune, Yoshua Bengio, and Hod Lipson.
\newblock How transferable are features in deep neural networks?
\newblock In {\em Advances in neural information processing systems}, pages
  3320--3328, 2014.

\bibitem{yue2015beyond}
Joe Yue-Hei~Ng, Matthew Hausknecht, Sudheendra Vijayanarasimhan, Oriol Vinyals,
  Rajat Monga, and George Toderici.
\newblock Beyond short snippets: Deep networks for video classification.
\newblock In {\em Proceedings of the IEEE conference on computer vision and
  pattern recognition}, pages 4694--4702, 2015.

\end{thebibliography}
}

\end{document}